# SMARtCARE: Privacy-Preserving Agentic AI Systems for Bounded-Autonomy Clinical Decision Support

Dr. Srini Ramaswamy
DNRS.ai USA
srini@computer.org

Dr. Deveeshree Nayak
University of Washington, USA
email: dnayak@uw.edu

***Abstract -*** *Long-context clinical AI systems can miss relevant patient history when prior admissions fall outside the active reasoning context. In ICU monitoring, this can cause early vital-sign drift to appear nonspecific even when it resembles a prior deterioration pattern. SMARtCARE addresses this gap through a four-state clinical decision-support architecture: Stable, Meta-cognitive, Assisted, and Regulated (Revoked). Rather than automatically retrieving prior records, SMARtCARE uses a lossy six-channel fingerprint of the patient's prior trajectory. When current drift matches that fingerprint and the prior record is absent from context, the system raises a Meta-cognitive escalation for clinician review; full retrieval occurs only through clinician action in the Assisted state. A patient-identity guard is designed to enforce correct attribution across data loading, logging, and audit layers. Evaluation combines a synthetic Monte Carlo study that validates the state-transition logic and estimator stability, not clinical performance, with real-data runs on both the MIMIC-III and MIMIC-IV Clinical Database Demos. On MIMIC-III, one prior-pattern recurrence was identified among 14 two-admission patients; on MIMIC-IV, the same pipeline produced no fingerprint matches among 9 two-admission patients, which illustrates a key limitation of a fixed canonical pattern library. Across both runs all logged decisions were fully traceable and correctly attributed. The results support SMARtCARE as a traceable, privacy-aware mechanism for surfacing middle-context risk; they are not a clinical efficacy claim.*

## I. INTRODUCTION

Clinical decision-support systems increasingly use AI models that maintain context across patient encounters. In ICU monitoring, this creates a practical limitation: the system's active reasoning context is finite. Prior admission may be documented in the electronic health record but absent from the information currently available to the monitoring agent. When this happens, a patient-specific deterioration pattern may not be considered during interpretation of new vital-sign drift. We call this condition middle-context loss: clinically relevant prior information exists but is in neither the admission baseline nor the recent observation stream. It is the clinical form of the "lost in the middle" problem in long-context models [1], with a sharper consequence - a prior deterioration trajectory may be unavailable exactly when current drift first becomes meaningful. Preloading prior records is not a safe fix: it expands inference-time exposure of protected information, and merely adding context does not ensure reliable use [1]. The safer alternative is explicit recognition of when context is insufficient, rather than more autonomous retrieval.

SMARtCARE is proposed as an architecture for handling this condition. The system stores a minimized representation of prior admission trajectory, compares current drift against that patient-specific fingerprint, and escalates to a clinician when a recurrence signal appears but the supporting prior record is not in context. The system does not make an autonomous clinical decision based on the fingerprint alone.

This paper makes four key contributions: (i) it defines middle-context loss as a decision-support failure mode in ICU monitoring; (ii) it introduces a four-state SMARtCARE architecture, instantiating a state-based autonomy framework [15], for surfacing this condition as a traceable escalation; (iii) it proposes a lossy six-channel trajectory fingerprint that supports recurrence detection while limiting prior-record exposure; and (iv) it evaluates the architecture using separated synthetic and real-data tracks - the latter on both the MIMIC-III and MIMIC-IV demos - with explicit limits on clinical interpretation.

The rest of this paper is organized as follows: Section II reviews related work and outlines the challenges of middle-context loss and agentic AI in clinical settings. Section III introduces the core SMARtCARE architecture, detailing its four bounded-autonomy operating states: Stable (S), Meta-cognitive (M), Assisted (A), and Regulated (Rt). Section IV details the simulation setup and results, evaluating the architecture through both synthetic state-logic validation and real-data ICU case studies. Finally, Section V concludes the paper and discusses future directions for safe, domain-grounded agentic AI in healthcare.

## II. RELATED WORK

The narrow problem SMARtCARE addresses is: how can a decision-support system recognize a patient-specific recurrence signal when the supporting prior admission is *not* in active context, without autonomously retrieving the full record? Four bodies of work bear on this question.

### *1. Context loss and its remedies*

Unlike SMARtCARE existing remedies for context loss in long-context models all assume the missing content will be loaded: Liu et al. [1] documented the mid-sequence degradation problem; Vaswani et al. [2] traced it to self-attention's lack of positional guarantee; and the dominant engineering responses - RAG reordering, mid-chunk compression, redundancy filtering, BriefContext map-reduce [23], TIMER temporal instruction tuning [21], and TrajOnco time-aware chunking with a long-term-memory module [22]; all reduce positional bias within a loaded prompt. None of them address the prior case where loading the record is itself the governance risk. SMARtCARE is positioned orthogonally: it treats *absence* of prior context as a first-class state and escalates, deferring retrieval to explicit clinician action. These methods are complementary to SMARtCARE, applicable once a system has been authorized to load the record; the contribution here is the layer below that decision. In [24], the authors present a structured framework for temporal annotation of clinical events using retrieval and LLM, with medically grounded prompting and consistency checks on the MIMIC-IV-Ext-22MCTS dataset featuring over 22 million clinical events. SMARtCARE is positioned orthogonally: rather than loading the history and attending to it better, it treats *absence* of prior context as a first-class state and escalates - the relevant default when loading the record is itself the governance risk. These methods are therefore complementary, applicable once a system has chosen (and is permitted) to load the record.

**Figure 1:** Middle-context loss and the fingerprint mechanism: primacy and recency zones of the current admission, the off-axis forgotten middle (prior admission), and the cosine match (0.841 ≥ 0.70) that raises the Meta-cognitive flag without auto-retrieval.

*2. Clinical early-warning and decision support*

Sepsis-3 [6] and SOFA [7] define the physiology against which deterioration models are benchmarked; TREWScore [8] showed EHR time-series models can predict septic shock hours early but are sensitive to population and sampling. Wong et al. [3] found a widely deployed proprietary sepsis model produced sensitivity and false-alert rates outside acceptable bounds in real use - the unresolved alert-fatigue/missed-detection tension. Sutton et al. [4] catalogued both the benefits and the alert-fatigue and override risks of decision-support systems, and Topol [5] and Rajkomar et al. [9] argued for augmentation over replacement. SMARtCARE occupies the intermediate regime these systems underserve - drift below a hard threshold but meaningful against a patient's own prior trajectory - and makes human review a logged, triggered condition (the A state) rather than a convention.

*3. Benchmarks and data infrastructure*

MIMIC-III [12], via PhysioNet [13], with the multitask benchmark suite of Harutyunyan et al. [18], was the standard ICU testbed; documented calibration drift and demographic bias on external validation [19] motivated the shift to MIMIC-IV [16] (>380,000 admissions, modular schema linking vitals, text, imaging, and ED logs). The DiReCT benchmark [20] adds MIMIC-IV multimodal evaluation with AUROC/AUPRC, demographic-fairness metrics, and SHAP interpretability. SMARtCARE is a routing/escalation mechanism, not a predictive model scored on these tasks; we use both demos for cross-version pipeline validation and flag DiReCT-style fairness and interpretability evaluation as prerequisites for any future predictive extension.

**S: Stable monitoring (Contextualize)**
|→ Sustained drift + patient-specific prior match + prior not in context → M
|→ Hard threshold breach → Rt
**M: Missing-context escalation (Assess)**
|→ Clinician pickup → A
**A: Clinician review (Refer)**
|→ Confirmed/dismissed → S
|→ Hard failure found → Rt
**Rt: Hard-threshold remediation (Escalate)**
|→ Resolved → S

**Figure 2.** SMARtCARE: Enabling Governed Intelligence

*4. Privacy and data minimization*

HIPAA Safe-Harbor [10] governs de-identification at rest but not inference-time context loading; pre-loading a full prior record expands the PHI exposure surface even on de-identified data. Differential privacy [11] offers a formal leakage framework whose application to clinical agents is still emerging. SMARtCARE's fingerprint is a representation-level instance of data minimization - a six-number vector sufficient for recurrence detection, with the full record accessed only on explicit, per-patient clinician action.

Taken together, no prior system combines state-based autonomy, a minimized trajectory fingerprint for recurrence detection, and a mandatory audit layer into a single architecture for absent-prior-context case; this combination is the contribution presented in this paper.

## III. THE SMARTCARE ARCHITECTURE

### *A. State-Based Autonomy in Clinical Monitoring*

The SMARt framework [15] motivates organizing clinical AI behavior around an explicit autonomy state rather than a single always-on inference mode. A context-bounded agent in the Stable state - comparing current vitals to a current baseline - cannot recognize that a signal uninterpretable in the present context would become interpretable against prior-stay data; lacking an intermediate state, it must either ignore the signal (missed detection) or auto-retrieve the record (PHI exposure). In OODA terms [15] this is an Orient-phase failure: interpretation requires prior-stay context absent from working memory. SMARtCARE adds that intermediate state - Meta-cognitive - which flags the condition and routes it to a clinician instead of choosing between the two unsafe defaults. Figure 1 illustrates the three memory zones - primacy (admission baseline), recency (live stream), and the forgotten middle (prior admission) - and the fingerprint-matching step that drives the transition.

### B. The Four States

SMARtCARE is organized as a four-state decision-support architecture separating routine monitoring, missing-context escalation, clinician review, and hard-threshold remediation (Regulated state). The states (Contextualize → Assess → Refer → Escalate), separate routine monitoring, missing-context escalation, clinician review, and hard-threshold intervention. The four lifecycle stages map to the Stable, Meta-cognitive, Assisted, and Regulated states respectively. Figure 2 shows the states and the transitions between them. (i) Stable (S) -Routine Monitoring (Contextualize): The conservative, default operating mode. It tracks vitals against an initial 24-hour baseline, ignoring transient noise. It remains in 'S' unless a hard clinical threshold is breached or a sustained drift matches a known prior trajectory. (ii) Meta-cognitive (M) - Missing-Context Escalation (Assess): The architecture's core safety mechanism. It triggers only when three conditions align: a sustained drift occurs, it matches a prior admission pattern,

**Table 1.** Monte Carlo results, 30 replications × 500 admissions, synthetic only

| Metric | Mean | SD | 95% CI low | 95% CI high |
|---|---|---|---|---|
| % ICU time in Stable (S) | 99.54 | 0.04 | 99.52 | 99.55 |
| % ICU time in Meta-cognitive (M) | 0.31 | 0.03 | 0.3 | 0.32 |
| % ICU time in Assisted (A) | 0.09 | 0.01 | 0.08 | 0.09 |
| % ICU time in Remediation/Truth (Rt) | 0.07 | 0.01 | 0.06 | 0.07 |
| Detection yield (confirmed / M escalations) | 0.81 | 0.04 | 0.79 | 0.82 |
| False-positive rate (dismissed / M escalations) | 0.19 | 0.04 | 0.18 | 0.2 |
| Forgotten-middle escalations per 100 admissions | 26.26 | 2.31 | 25.43 | 27.09 |
| Forgotten-middle retrieved per 100 admissions | 21.19 | 2.22 | 20.39 | 21.98 |
| Rt hard-failure detections per 100 admissions | 11.53 | 1.55 | 10.97 | 12.08 |

*and* that prior record is missing from the active context. It does not alter care or autonomously pull records; instead, it converts the missing context into a structured, auditable alert (containing the signal, IDs, and prior pointer). (iii) Assisted (A) - Clinician Review (Refer): Triggered when a clinician responds to an 'M' escalation. The clinician reviews the historical record and decides the outcome: escalate monitoring, dismiss the alert (return to 'S'), or declare a critical failure (transition to 'Rt'). This closes the context gap and prevents duplicate alerts for the same signal. (iv) Regulated (Rt) - Hard-Threshold Intervention (Escalate): The ultimate safety override. Triggered strictly by verifiable, life-threatening data (e.g., severe hypotension, critical potassium, allergy conflicts). It bypasses predictive pattern-matching entirely, immediately revoking autonomous handling and transferring full control to the care team.

### C. The Four-Agent Pipeline

The architecture decomposes into four agents with non-overlapping responsibilities so that no single agent combines sensing, decision, execution, and audit. The Baseline Curator computes an immutable per-admission baseline (channel-wise median and standard deviation for seven vitals, forward-fill imputation for gaps up to six hours, and a confidence classification of full ≥90% / sparse 50–90% / very sparse <50%); a sparse baseline raises the M-escalation threshold. The Pattern Detector computes per-channel z-scores, confirms sustained drift (≥2 consecutive readings >2 SD from baseline median over ≥2 hours), and only then computes cosine similarity between the observed six-channel drift vector and the patient's own fingerprint - a generic library match alone is insufficient. The Escalation Manager owns all state transitions and Rt threshold checks, manages the A-state review, closes the context gap on resolution, and removes a resolved medication contraindication from the working list. The Outcome Auditor is the single logging authority and rejects any event missing required lineage (patient/admission identifiers, triggering rule, vital values with MIMIC source table and itemid, baseline values, computed metrics, pattern-match detail).

**Table 2.** Verified itemids-to-channel mapping for MIMIC-III and MIMIC-IV

| Channel | MIMIC-III itemids (CareVue / MetaVision) | MIMIC-IV itemids (MetaVision only) | Verified label |
|---|---|---|---|
| hr | 211, 220045 | 220045 | Heart Rate |
| sbp | 51, 442, 455, 220179, 220050 | 220179, 220050 | Arterial / NBP Systolic BP |
| dbp | 8368, 8440, 8441, 8555, 220180, 220051 | 220180, 220051 | Arterial / NBP Diastolic BP |
| map | 456, 52, 6702, 443, 220052, 220181 | 220052, 220181 | Arterial / NBP Mean BP |
| temp | 223761, 678 (F); 223762, 676 (C) | 223761, 678 (F); 223762, 676 (C) | Temperature (converted by itemid identity) |
| rr | 615, 618, 220210, 224690 | 220210, 224690 | Respiratory Rate |
| o2_sat | 646, 220277 | 220277 | SpO2 / O2 Saturation pulseoxymetry |
| lactate | 50813 | 50813 | Lactate (Blood Gas) |
| creatinine | 50912 | 50912 | Creatinine (Chemistry) |
| potassium | 50822, 50971 | 50822, 50971 | Potassium (Blood Gas + Chemistry) |
| wbc | 51300, 51301 | 51300, 51301 | White Blood Cells (Hematology) |
| platelets | 51265 | 51265 | Platelet Count |
| glucose | 50809, 50931 | 50809, 50931 | Glucose (Blood Gas + Chemistry) |
| bun | 51006 | 51006 | Urea Nitrogen ("BUN", Chemistry) |

The Pattern Detector's per-patient work is bounded and predictable: z-score computation is $O(c)$ per reading where $c = 6$ channels; cosine similarity between the drift vector and the fingerprint library is $O(c \times |L|)$ where $|L|$ is the library size (currently 4 patterns). The only per-patient lookup is a single fingerprint retrieval keyed by patient identifier; no pairwise comparison across patients is required. In a 20-bed ICU with vitals charted every 15 minutes, the detector executes roughly 1,920 z-score evaluations and up to 7,680 cosine operations per hour across the unit - well within single-machine capacity. Alert volume is self-limiting by design: M escalations require three simultaneous conditions (sustained drift, fingerprint match, and prior record absent from context), and once a clinician reviews a prior pattern in the A state, that pattern is closed for the remainder of the stay, preventing repeat alerts for the same signal. False-positive load therefore does not grow linearly with cohort size; it is bounded by the number of distinct fingerprinted prior stays per patient. At population scale (hundreds of beds), the stateless per-patient design supports straightforward horizontal partitioning by patient. Full throughput characterization under realistic ICU data rates is identified as future work.

### D. The Fingerprint as a Task-Sufficient Minimal Representation

The fingerprint is constructed from the prior admission's own first-48-hour vital trajectory. For each of six channels (HR, SBP, MAP, temperature, RR, lactate) the algorithm takes the median of the earliest quartile and the median of the latest quartile of the 48-hour window, then divides the early-to-late difference by a reference standard deviation (8 bpm, 12 mmHg, 8 mmHg, 0.5 °C, 3 breaths/min, 1.0 mmol/L). These reference values are population-level estimates of within-patient hourly variability, chosen to make the delta vector comparable across patients; they are not derived from the individual patient's own variance. This is a deliberate comparability choice with a known cost: for patients whose chronic physiology (e.g., end-stage renal disease, severe COPD) departs markedly from population norms, population-level normalization can dampen sensitivity. Per-patient adaptive normalization is identified as future work. The six-number vector is matched against a small library of canonical patterns by cosine similarity; a label is assigned only if maximum channel movement exceeds 1.5 reference SD and cosine similarity to a labelled pattern exceeds 0.6. A prior stay that was stable or moved incoherently receives a null label, and no M escalation can fire for it.

***The representation is deliberately lossy:*** it preserves the direction and approximate magnitude of physiological change and discards absolute values, timestamps, medications, notes, free text, and identifiers. The prior admission identifier is retained only as a retrieval pointer for the A state. The data-minimization claim is bounded - the fingerprint is the minimal representation sufficient for recurrence detection, not a lossless summary, not a complete clinical representation, and not a de-identification mechanism.

***Positioning against conceptual baselines:*** Three natural comparators clarify what SMARtCARE adds. A *primacy-and-recency-only agent*, the implicit baseline in most deployed systems, would assess the anchor-case drift (MAP 79 mmHg, HR 99, no hard threshold breached) as nonspecific and not escalate; SMARtCARE raised an M flag and surfaced the prior sepsis pattern. A *full-history retrieval agent* would load every prior admission at inference time, resolving the context gap at the cost of expanding PHI exposure and increasing context length for every patient regardless of whether a recurrence signal exists. A *threshold-only agent* (analogous to NEWS2 or SOFA thresholds applied directly) would not fire on the anchor case at all; it only acts when hard limits are crossed, which is exactly the regime the M-state is designed to precede. Formal quantitative comparison against these baselines requires a credentialed cohort large enough to estimate sensitivity and specificity, a prerequisite stated later. The present evaluation establishes that the pipeline executes correctly and that the M state fires in at least one real case where all three comparators would remain silent.

### *E. Patient-Identity Integrity*

A recurrence flag, retrieved history, or modified monitoring plan must be associated with the patient for whom it was computed. SMARtCARE enforces this at three points. At data-load time, the loader builds an authoritative admission-to-patient mapping, verifies injectivity (each admission belongs to exactly one patient), and confirms that every prior-to-current admission link connects admissions of the same patient - any violation aborts the load. At logging time, the Auditor requires a patient identifier on every event.

**Table 3.** Combined real-data results, MIMIC-III vs MIMIC-IV

| Metric | MIMIC-III Demo v1.4 (N=14) | MIMIC-IV Demo v2.2 (N=9) |
|---|---|---|
| Total ICU-hours logged | 1163 | 1194 |
| Mean ICU stay length (h) | 83 | 133 |
| % time in Stable (S) | 94.9 | 96 |
| % time in Meta-cognitive (M) | 0.09 | 0 |
| % time in Assisted (A) | 0.03 | 0 |
| % time in Remediation/Truth (Rt) | 4.99 | 4 |
| Prior stays fingerprinted | 3/14 (21%) | 0/9 (0%) |
| M escalations | 1 | 0 |
| M escalations confirmed | 1 | n/a |
| M escalations dismissed (false positives) | 0 | n/a |
| Rt hard-failure detections | 116 | 95 |
| Rt detections per 100 ICU-hours | 10 | 7.9 |
| Total logged events | 235 | 190 |
| Events fully traceable (%) | 100 | 100 |
| Events identity-clean (%) | 100 | 100 |

At audit time, the metrics report verifies that each logged event's patient identifier matches the load-time mapping. The underlying MIMIC data satisfies HIPAA Safe-Harbor de-identification (no name, MRN, SSN, or address fields; dates shifted by a per-patient offset; surrogate-integer identity key), verified by inspection of all table headers in each demo release. In the implementation evaluated here, these checks prevented cross-patient attribution at both the data and application layers; we report this as a property of the present implementation rather than a general guarantee.

## IV. SIMULATION SETUP AND RESULTS

### *A. Data Sources and Evaluation Scope*

Three data sources are used and kept strictly separate; every record carries a `source` tag that persists through all metrics. Synthetic data (`SYNTHETIC_FIXTURE`) is used only to characterize state-machine behavior and estimator stability under stated assumptions; it makes no clinical claim. MIMIC-III Clinical Database Demo v1.4 [12], [14] (ODbL, `MIMIC_III_DEMO`) and MIMIC-IV Clinical Database Demo v2.2 [16], [17] (`MIMIC_IV_DEMO`) provide two independent real-data tracks. Both demos are openly licensed and require no credentialed PhysioNet access, CITI training, or data-use agreement, which is why they are used here; both are too small for population inference, and the real-data results are presented as feasibility and boundary-condition demonstrations rather than statistical validation. Using openly licensed demos also keeps the pipeline and its artifacts reproducible without access barriers, which is intended to support independent replication and collaborative extension to credentialed cohorts. The simulation environment - the four-agent pipeline, the MIMIC-III and MIMIC-IV loaders, the Monte Carlo harness, and the metrics and audit tooling - was implemented and executed with the assistance of Anthropic's Claude Code; all generated artifacts (code, figures, per-table CSV summaries, and decision logs) are retained for reproducibility.

**Table 4.** Prior-stay fingerprint results across both demo cohorts

| Dataset | Cohort N | Priors with fingerprint | Patterns identified | Re-exhibited on current stay | M escalations |
|---|---|---|---|---|---|
| MIMIC-III Demo v1.4 | 14 | 3 (21%) | 2 sepsis_early, 1 cardiac_decompensation | 1 (adm 170883) | 1 |
| MIMIC-IV Demo v2.2 | 9 | 0 (0%) | none | 0 | 0 |

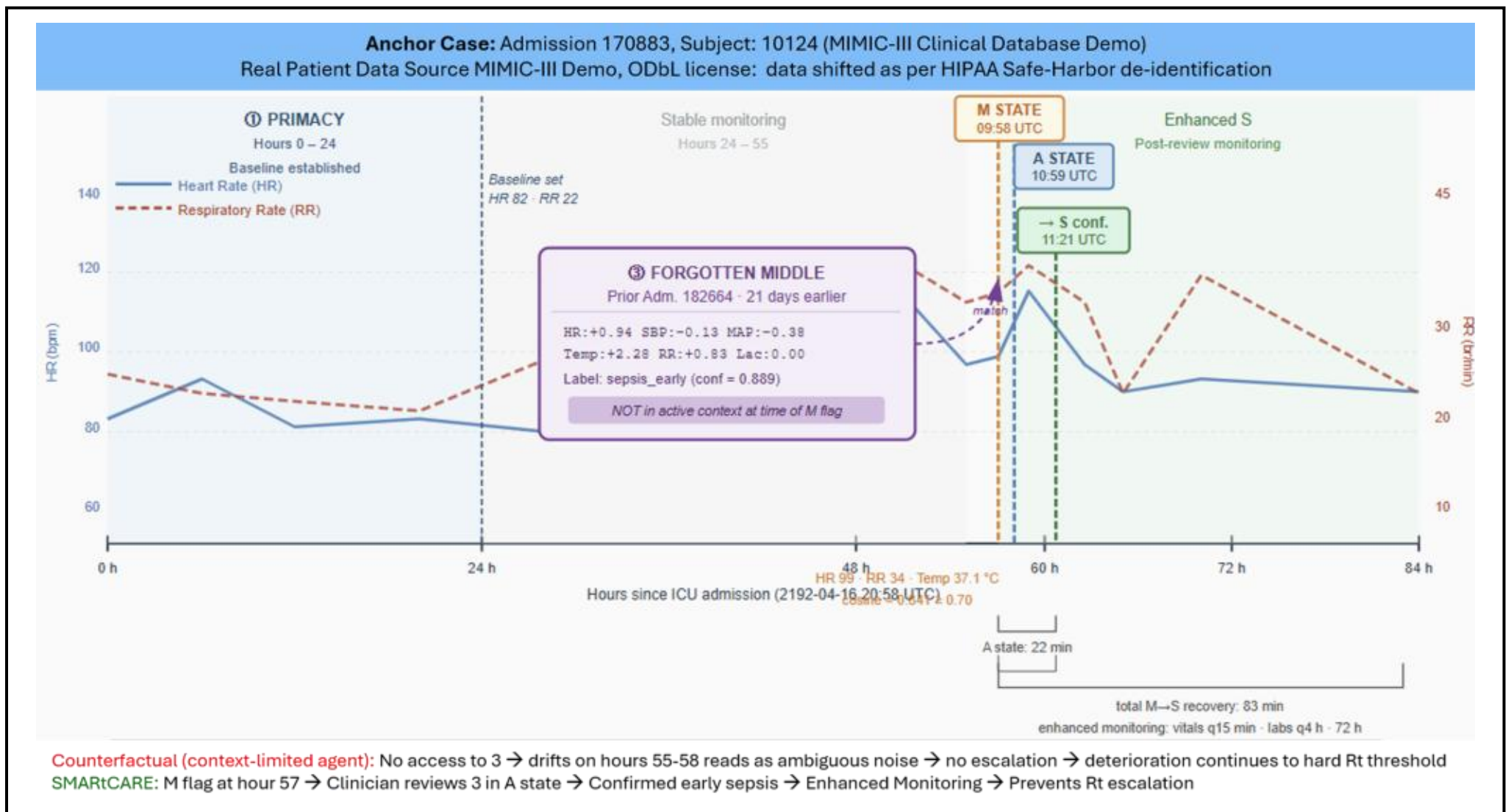

**Figure 3:** Anchor-case timeline (MIMIC-III admission 170883): HR and RR traces across the 84-hour stay with the M (09:58 UTC), A (10:59 UTC), and confirmed-S (11:21 UTC) markers

A note on the role of the synthetic study: because the synthetic generator produces data consistent with the same archetype and pattern-library assumptions the pipeline encodes, the Monte Carlo study is tautological-by-construction with respect to clinical performance. It can establish that the state-transition logic routes data correctly and that the resulting estimators are stable under resampling; it cannot provide evidence about real-world sensitivity or false-alert rates. We report it only for the former purpose.

### B. Verified Variable Mapping

All vital and laboratory itemids were verified against each demo's dictionary tables (`D_ITEMS`/`D_LABITEMS`) and confirmed present in `CHARTEVENTS`/`LABEVENTS` before analysis (Table 2). MIMIC-III mixes CareVue and MetaVision identifiers; MIMIC-IV is MetaVision-only, so CareVue-specific ids (e.g., HR 211, SBP 51) are absent there. Temperature is converted to Celsius by itemid identity (Fahrenheit set {223761, 678}), never by value heuristic; unrecognized temperature itemids are discarded. Blood-pressure channels pool invasive arterial, non-invasive, and manual sources (documented simplification). Prior diagnoses are resolved to titles via the ICD dictionary (ICD-9 for MIMIC-III, ICD-10 for MIMIC-IV).

### C. Synthetic Monte Carlo: State-Logic and Estimator Stability

Thirty independent replications of 500 admissions each (15,000 draws, Table 1) were drawn from a four-archetype mixture (≈58% stable, 25% recurrence, 10% hard-failure, 7% prior-in-context), with independent seeds for cohort and synthetic clinician. It reports means and 95% confidence intervals; the narrow intervals indicate estimator stability under resampling. The detection yield (0.81) and false-positive rate (0.19) describe how the routing logic partitions M escalations under the synthetic-clinician policy, and the forgotten-middle rate (≈21.2 per 100 admissions, [20.4–22.0]) counts escalations a primacy-and-recency-only agent would not raise. As noted these are properties of the state logic and the generative assumptions, not clinical effect sizes.

### D. Real-Data Evaluation: MIMIC-III and MIMIC-IV Demos

Both real cohorts use the same definition: patients with ≥2 ICU admissions where both the prior and current admission have an ICU stay carrying vital chartevents. Admissions without ICU chartevents (e.g., general-ward encounters) are excluded rather than admitted with empty baselines; in the MIMIC-IV demo this filter reduces 48 two-hospital-admission subjects to 9 with usable ICU vitals on both stays. Under this definition MIMIC-III yields 14 patients and MIMIC-IV yields 9. For each patient the fingerprinter was applied to the most recent prior admission (Table 4). As summarized in the table, on MIMIC-III 3 of 14 prior stays produced a recognizable fingerprint (two early-sepsis, one cardiac decompensation) and the other 11 returned null; one of the three re-exhibited the prior pattern on the current stay, producing the single M escalation in that cohort (the anchor case - Figure 3). On MIMIC-IV, 0 of 9 prior stays produced a fingerprint under the current four-pattern library, and therefore no M escalation occurred. This is not an artifact of sparse data: the most active MIMIC-IV prior stay (admission 23559586) showed a large excursion (maximum standardized movement 4.0 reference SD) but a falling-lactate trajectory (−4.0 SD) consistent with recovery from a prior sepsis episode - directionally opposite the canonical onset shapes, and therefore correctly left unlabeled. This illustrates the limitation the fixed canonical library imposes: it represents a small set of syndromic onset shapes and may not capture idiosyncratic or recovery-direction trajectories, which motivates a learned pattern library. The contrast across both cohorts is shown in Figure 4, the divergence in fingerprint and M-escalation counts appears to reflect pattern-library breadth, while the library-independent Rt layer (Table 4) behaved comparably on both datasets, indicating the hard-threshold safety path operates on real vitals from either source.

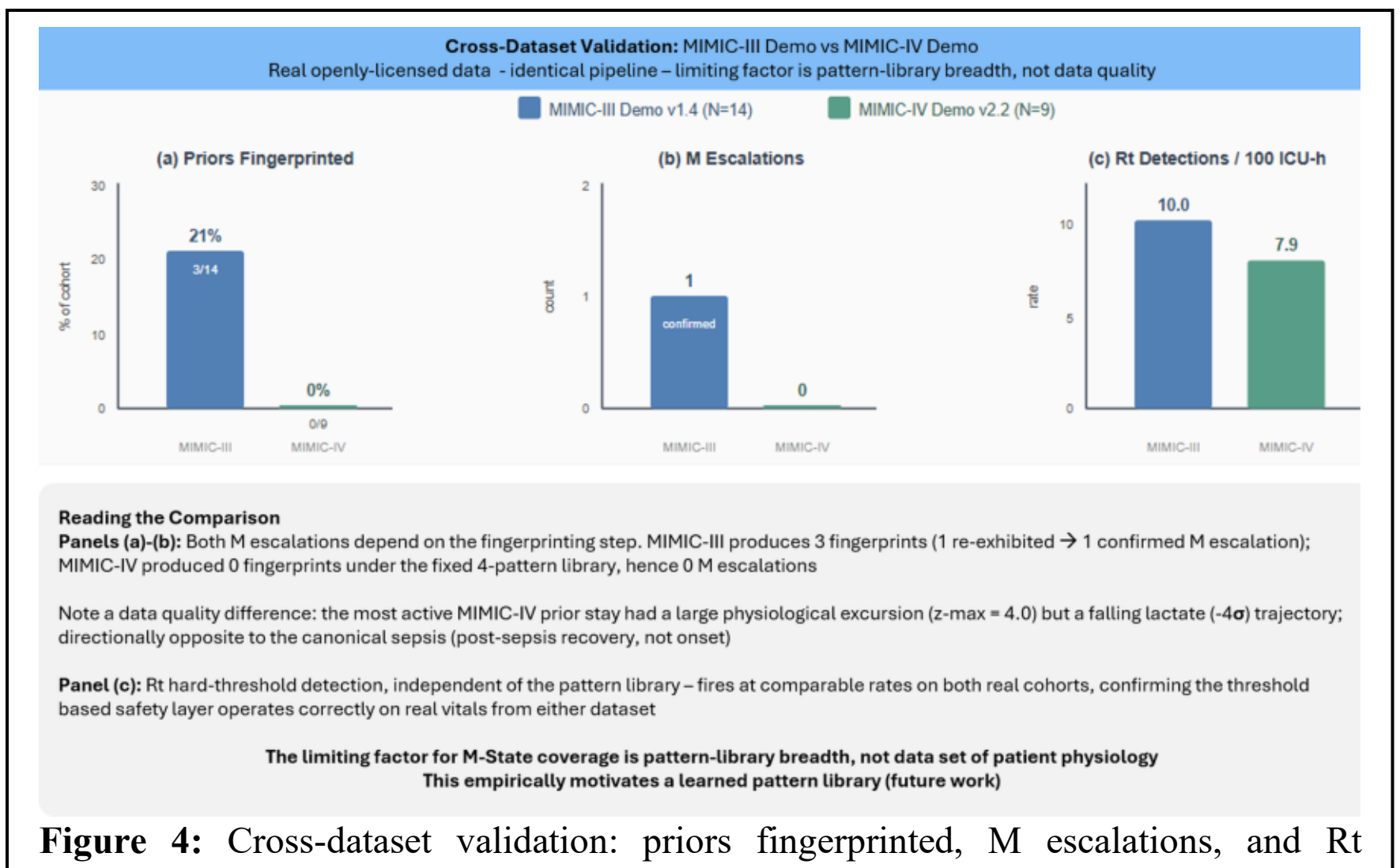

**Figure 4:** Cross-dataset validation: priors fingerprinted, M escalations, and Rt detections per 100 ICU-hours for MIMIC-III vs MIMIC-IV.

### *E. Anchor Case: Admission 170883, Subject 10124 (MIMIC-III)*

The prior admission (182664), 21 days earlier, was fingerprinted as early-sepsis with cosine confidence 0.889 (delta vector HR +0.94, SBP −0.13, MAP −0.38, Temp +2.28, RR +0.83, Lactate 0.0 in reference SD units; prior diagnoses included CHF, chronic and acute kidney disease, and atrial fibrillation). On day three at 09:58 UTC the Pattern Detector confirmed a sustained two-hour drift (HR 99 bpm, RR 34 breaths/min, Temp 37.1 °C); cosine similarity to the prior fingerprint was 0.841 (> 0.70), and the prior record was absent from context. The system entered M and logged an escalation with the matched pattern, confidence, and a retrieval pointer. The A state began at 10:59 UTC; the simulated clinician retrieved admission 182664, confirmed the risk at 11:21 UTC (A-state 22 minutes, total M→S recovery 83 minutes), and the plan was upgraded to enhanced monitoring. Figure 3 plots the vital-sign timeline and the state-transition events for this admission. Under the simulation model, a primacy-and-recency-only system would assess the same drift below hard-threshold criteria (MAP 79 mmHg > 55; HR 99 < 140) and not escalate. Whether this difference would change a real clinical outcome is outside what the simulation can establish; the anchor case illustrates mechanism, not an incidence rate.

### *F. Traceability and Identity Audit*

Across the two real runs, (results in Table 3) 425 state-transition events were logged (235 MIMIC-III, 190 MIMIC-IV). All carried complete data lineage (source tag, MIMIC table, itemid, charttime) and all passed the patient-identity audit with zero violations. Zero prior records were retrieved autonomously in either run; the single retrieval (MIMIC-III) occurred at the A state on clinician action for the one confirmed escalation.

### *G. Limitations*

Implementation parameters: Table 5 summarizes the key thresholds governing state transitions. The Rt layer checks are evaluated on every vital update; M-layer checks require the sustained-drift condition to hold for ≥2 consecutive readings over ≥2 hours before cosine similarity is computed, preventing transient spikes from triggering escalation. Fingerprint cosine threshold (0.70) and pattern-library confidence threshold (0.60) were set by inspection on the synthetic cohort and held fixed for the real-data runs — no tuning was performed on real data.

Audit record structure. Every logged event carries: patient_id, admission_id, source (MIMIC table + itemid), charttime, state_from, state_to, trigger_rule, vital_values (all 6 channels), baseline_values, z_scores, pattern_match (label + cosine), and prior_admission_pointer. A representative M-escalation record for the anchor case is shown in Figure 5 below.

**Table 5. State-transition thresholds**

| Parameter | Value | Governs |
|---|---|---|
| Sustained drift window | ? 2 consecutive readings, ? 2 h | M entry condition |
| Drift magnitude (z-score) | > 2.0 SD from baseline median | Per-channel trigger |
| Cosine similarity (M flag) | ? 0.70 | Drift vs. fingerprint match |
| Fingerprint library confidence | ? 0.60 cosine + max channel > 1.5 SD | Prior stay labelling |
| Baseline confidence: full | ? 90% vital coverage | Normal M threshold |
| Baseline confidence: sparse | 50–90% vital coverage | Raised M threshold |
| Rt: hypotension | MAP < 55 mmHg (sustained) | Hard-threshold path |
| Rt: tachycardia | HR > 140 bpm | Hard-threshold path |
| Rt: hyperthermia | Temp > 39.5 °C | Hard-threshold path |
| Rt: critical potassium | K? < 2.5 or > 6.5 mEq/L (on relevant meds) | Hard-threshold path |
| Gap imputation limit | ? 6 h forward-fill | Baseline Curator |

Both real cohorts (N = 14, N = 9; non-overlapping) are far too small for population inference; all real-data percentages are descriptive of these specific patients. The synthetic results are valid only under the stated archetype and pattern-library assumptions and are tautological-by-construction with respect to clinical performance. The synthetic clinician is a deterministic policy, not a study of human decision-making. Two design limitations include: (i) population-level reference SD can dampen sensitivity for chronically abnormal physiology, and (ii) the fixed canonical pattern library cannot represent idiosyncratic or recovery-direction trajectories (the MIMIC-IV run, with 0/9 fingerprints despite real physiological movement, is direct evidence). Full-cohort validation on credentialed MIMIC-IV (available through PhysioNet with CITI training and a data-use agreement), a learned pattern library, per-patient normalization, DiReCT-style fairness and interpretability evaluation, and a prospective study of clinician response are required before any clinical-utility claim.

```
{
  "patient_id": 10124,  "admission_id": 170883,
  "source": "MIMIC_III_DEMO / CHARTEVENTS",
  "charttime": "2192-04-19T09:58:00Z",
  "state_from": "S",  "state_to": "M",
  "trigger_rule": "sustained_drift_fingerprint_match",
  "vital_values":  {"HR":99,"SBP":118,"MAP":79,"Temp":37.1,"RR":34,"Lactate":null},
  "baseline_values":{"HR":82,"SBP":124,"MAP":86,"Temp":36.7,"RR":22,"Lactate":null},
  "z_scores":       {"HR":2.1,"SBP":-0.5,"MAP":-0.9,"Temp":0.8,"RR":6.0},
  "pattern_match":  {"label":"sepsis_early","cosine":0.841,"library_cosine":0.889},
  "prior_admission_pointer": 182664,
  "autonomous_retrieval": false
}
```

**Figure 5. Representative M-State Escalation**

## V. CONCLUSIONS

SMARtCARE makes three primary contributions. First, it defines *middle-context loss*, the condition in which a clinically relevant prior admission exists in the record but is absent from an AI agent's active reasoning context; as a distinct decision-support failure mode requiring an architectural, not just a retrieval-based, response. Second, it instantiates a four-state bounded-autonomy architecture (Stable, Meta-cognitive, Assisted, Regulated/Revoked) in which absence of prior context is a first-class, logged, reviewable state rather than a silent gap: when current drift matches a patient-specific prior trajectory and the supporting record is not loaded, the system escalates to a clinician rather than auto-retrieving or ignoring the signal. Third, it demonstrates that a lossy six-channel trajectory fingerprint is sufficient to detect recurrence signals while limiting inference-time PHI exposure, and that a mandatory audit layer can enforce complete data lineage and correct patient attribution across a full pipeline run.

The synthetic Monte Carlo study validates state-transition logic and estimator stability; it does not speak to clinical performance. The MIMIC-III demo run confirmed pipeline operation with one real prior-pattern recurrence; the identical MIMIC-IV run produced no fingerprint matches, empirically demonstrating the limitation of a fixed canonical library rather than a data-quality problem. These results support SMARtCARE as a privacy-aware mechanism for surfacing missing-context risk. They do not establish clinical efficacy, generalizability, or deployment readiness.

SMARtCARE is not a deterioration predictor. It is a privacy-preserving escalation architecture that detects when patient-specific prior context may be needed, converts that condition into an auditable event, and routes retrieval to clinician action.

Future work falls into four tracks. (1) *Validation*: full credentialed MIMIC-IV with DiReCT-style fairness (demographic parity, equalized odds) and interpretability (SHAP) evaluation, and a learned pattern library replacing the fixed four-shape set. (2) *Normalization*: per-patient adaptive reference SD to restore sensitivity for chronically abnormal physiology (ESRD, COPD). (3) *Privacy formalization*: differential-privacy analysis of fingerprint leakage under reconstruction attacks. (4) *Human factors*: a prospective study of clinician response to M-state escalations, measuring time-to-review, alert acceptance rate, override rate, and downstream care-plan changes across at least three ICU settings. The study design should include a control arm (standard monitoring without M-state alerts), a washout period between arms to limit learning effects, and structured clinician feedback on alert clarity and workflow fit. Alert-fatigue risk, the primary concern at scale; cannot be assessed without this study; it is a prerequisite for any deployment claim.

***Acknowledgments:*** *The authors used AI-assisted coding tools for simulation. All outputs were reviewed and validated by the authors. The authors welcome collaboration on credentialed-cohort validation and prospective clinical study.*

## VI. REFERENCES

[1] N. F. Liu et al., "Lost in the Middle: How Language Models Use Long Contexts," *Trans. Assoc. Comp. Linguistics*, vol. 12, pp. 157–173, 2024. arXiv:2307.03172.

[2] A. Vaswani et al., "Attention Is All You Need," in *Adv. Neural Information Processing Systems*, vol. 30, 2017.

[3] A. Wong et al., "External Validation of a Widely Implemented Proprietary Sepsis Prediction Model in Hospitalized Patients," *JAMA Internal Medicine*, vol. 181, no. 8, pp. 1065–1070, 2021.

[4] R. T. Sutton et al., "An overview of clinical decision support systems: benefits, risks, and strategies for success," *npj Digital Medicine*, vol. 3, art. 17, 2020.

[5] E. J. Topol, "High-performance medicine: the convergence of human and artificial int.," *Nature Medicine*, vol. 25, no. 1, pp. 44–56, 2019.

[6] M. Singer et al., "The Third International Consensus Definitions for Sepsis and Septic Shock (Sepsis-3)," *JAMA*, vol. 315, no. 8, pp. 801–810, 2016.

[7] J.-L. Vincent et al., "The SOFA score to describe organ dysfunction/failure," *Intensive Care Medicine*, vol. 22, no. 7, pp. 707–710, 1996.

[8] K. E. Henry, D. N. Hager, P. J. Pronovost, and S. Saria, "A targeted real-time early warning score (TREWScore) for septic shock," *Science Translational Medicine*, vol. 7, no. 299, p. 299ra122, 2015.

[9] A. Rajkomar et al., "Scalable and accurate deep learning with electronic health records," *npj Digital Medicine*, vol. 1, art. 18, 2018.

[10] U.S. Dept. of Health and Human Services, "Standards for Privacy of Individually Identifiable Health Information; Final Rule," *45 C.F.R. 164.514(b) (Safe Harbor De-identification)*, 2000.

[11] C. Dwork, "Differential Privacy," in *Proc. ICALP*, LNCS vol. 4052, pp. 1–12, Springer, 2006.

[12] A. E. W. Johnson et al., "MIMIC-III, a freely accessible critical care database," *Scientific Data*, vol. 3, art. 160035, 2016.

[13] A. L. Goldberger et al., "PhysioBank, PhysioToolkit, and PhysioNet," *Circulation*, vol. 101, no. 23, pp. e215–e220, 2000.

[14] A. E. W. Johnson, T. J. Pollard, and R. G. Mark, "MIMIC-III Clinical Database Demo," v1.4, *PhysioNet*, 2019. doi:10.13026/C2HM2Q.

[15] S. Ramaswamy, "Intelligence as Managed Autonomy: Failure, Escalation, and Governance for Agentic AI Systems," *J. Intelligent & Robotic Systems,* to appear.

[16] A. E. W. Johnson et al., "MIMIC-IV, a freely accessible electronic health record dataset," *Scientific Data*, vol. 10, art. 1, 2023. doi:10.1038/s41597-022-01899-x.

[17] A. E. W. Johnson, L. Bulgarelli, T. J. Pollard, S. Horng, L. A. Celi, and R. G. Mark, "MIMIC-IV Clinical Database Demo," v2.2, *PhysioNe*t, 2023. doi:10.13026/dp1f-ex47.

[18] H. Harutyunyan, H. Khachatrian, D. C. Kale, G. Ver Steeg, and A. Galstyan, "Multitask learning and benchmarking with clinical time series data," *Scientific Data*, vol. 6, art. 96, 2019.

[19] E. Röösli, S. Bozkurt, and T. Hernandez-Boussard, "Peeking into a black box, the fairness and generalizability of a MIMIC-III benchmarking model," *Scientific Data*, vol. 9, art. 24, 2022.

[20] B. Wang, et al., "DiReCT: Diagnostic Reasoning for Clinical Notes via Large Language Models," v1.0.0, *PhysioNet*, 2024, arXiv:2408.01933.

[21] H. Cui, et.al. "TIMER: temporal instruction modeling and evaluation for longitudinal clinical records." *npj Digital Medicine* 8, no. 1, 2025: 577. https://doi.org/10.1038/s41746-025-01965-9. arXiv:2503.04176.

[22] S. Zeng, et al, "TrajOnco: A Multi-Agent Framework for Temporal Reasoning over Longitudinal EHR for Multi-Cancer Early Detection," arXiv:2604.10386, 2026.

[23] G. Zhang, et al, "Leveraging Long Context in Retrieval-Augmented Language Models for Medical Question Answering," *npj Digital Medicine*, vol. 8, art. 239, 2025. https://doi.org/10.1038/s41746-025-01651-w

[24] Wang, J., et al, "MIMIC-IV-Ext-22MCTS: A 22 Millions-Event Temporal Clinical Time-Series Dataset with Relative Timestamp", (version 1.0.0). *PhysioNet*. 2025 https://doi.org/10.13026/dkj6-r828